\documentclass[conference,compsoc]{IEEEtran}
\usepackage{booktabs} % For \toprule, \midrule, \bottomrule
\usepackage{makecell} % For \makecell
\usepackage{subcaption} % For \subfigure and \subcaption

\usepackage[linesnumbered,ruled,vlined]{algorithm2e} % For algorithm2e environment
\usepackage{amsmath} % For \text
\usepackage{multirow}
\usepackage{bm}
\usepackage{cite}
\usepackage{caption}
\usepackage[numbers]{natbib}
\usepackage{mathrsfs}
\usepackage[table]{xcolor}
\usepackage{url}
\usepackage{soul}

\definecolor{pink}{rgb}{1.0, 0.85, 0.95}

\ifCLASSINFOpdf
  \usepackage[pdftex]{graphicx}
\else
  \usepackage[dvips]{graphicx}
\fi
\begin{document}
\title{An Empirical Study of Accuracy-Robustness Tradeoff and\\
Training Efficiency in Self-Supervised Learning}

\author{\IEEEauthorblockN{Fatemeh Ghofrani}
\IEEEauthorblockA{College
of Engineering and Computing\\ University of South Carolina\\
Columbia, South Carolina, USA\\E-mail: ghofrani@email.sc.edu}
\and
\IEEEauthorblockN{Pooyan Jamshidi}
\IEEEauthorblockA{College
of Engineering and Computing\\ University of South Carolina\\
Columbia, South Carolina, USA\\ E-mail: pjamshid@cse.sc.edu.}
}


% compsoc conferences compared to traditional conferences), use this
%Telephone: (800) 555--1212, Fax: (888) 555--1212}


\maketitle

\begin{abstract}
Self-supervised learning (SSL) has significantly advanced in learning image representations, yet efficiency challenges persist, particularly under adversarial training. Many SSL methods require extensive training epochs to achieve convergence, a demand further amplified in adversarial settings. To address this inefficiency, we revisit the robust EMP-SSL framework, emphasizing the crucial role of increasing the number of crops per image instance to accelerate the learning process. Unlike conventional contrastive learning, robust EMP-SSL leverages multiple crops per image, integrates an invariance term and regularization, and significantly reduces the required training epochs, enhancing time efficiency. Additionally, robust EMP-SSL is evaluated using both standard linear classifiers and multi-patch embedding aggregation, providing new insights into SSL evaluation strategies. This paper investigates these methodological improvements with a focus on adversarial robustness, combining theoretical analysis with comprehensive experimental evaluation. Our results demonstrate that robust crop-based EMP-SSL, when evaluated with standard linear classifiers, not only accelerates convergence but also achieves a superior balance between clean accuracy and adversarial robustness, outperforming multi-crop embedding aggregation. Furthermore, we extend this approach by applying free adversarial training to the Multi-Crop Self-Supervised Learning algorithm, resulting in the Cost-Free Adversarial Multi-Crop Self-Supervised Learning (CF-AMC-SSL) method. This method shows the effectiveness of free adversarial training in self-supervised learning, particularly when the number of epochs is reduced. Our findings underscore the efficacy of CF-AMC-SSL in simultaneously improving clean accuracy and adversarial robustness within a reduced training time, offering promising avenues for practical applications of SSL methodologies. Our code is released at \textcolor{magenta}{\url{https://github.com/softsys4ai/CF-AMC-SSL}}. 
\end{abstract}


% \fi
\IEEEpeerreviewmaketitle

\section{Introduction}

In recent years, progress in self-supervised learning (SSL) \cite{balestriero2023cookbook} has produced representations that match or exceed those achieved by supervised learning in classification tasks \cite{chen2020simple, grill2020bootstrap}. This advancement has led to state-of-the-art performance in various applications (e.g., models like BERT and GPT-3) \cite{brown2020language, devlin2018bert}. 

A prominent method in the realm of Self-Supervised Learning (SSL), namely joint-embedding SSL \cite{bardes2021vicreg, chen2020simple, zbontar2021barlow}, is primarily concerned with the generation of uniform representations for image augmentations. As delineated by Wu \cite{wu2018unsupervised}, the principle of instance contrastive learning requires the training of dual networks to produce analogous embeddings for disparate views of an identical image within a joint embedding framework. Originating from the Siamese network architecture \cite{bromley1993signature}, these methodologies deal with a fundamental obstacle, colloquially termed `collapse,' where all representations converge to uniformity, thereby disregarding input heterogeneity. To counteract this collapse phenomenon, two principal strategies have been proposed: contrastive and information maximization. Contrastive learning methodologies aim to identify dissimilar samples employing current branches or memory banks \cite{chen2020simple, he2020momentum}. On the contrary, noncontrastive methodologies such as TCR \cite{li2022neural}, Barlow Twins \cite{zbontar2021barlow}, and VICReg \cite{bardes2021vicreg} exploit covariance regularization as a means to avert the collapse of representations. Simultaneously, the technique of Swapping Assignments between Views (SwAV) \cite{caron2020unsupervised} delves into the application of multi-crop techniques in self-supervised learning, favoring a combination of views with diverse resolutions over a pair of full-resolution views. This groundbreaking research underscores the potential for enhanced SSL performance via multi-view augmentation, thereby offering a potent strategy to combat the issue of collapsing representations.

Despite the considerable potential displayed by self-supervised learning to derive effective representations from extensive unlabeled data, its vulnerability to adversarial attacks remains a substantial concern within the field \cite{ghofrani2023rethinking,kim2020adversarial,wahed2022adversarial}.  In response to these attacks, adversarial training is one of the most reliable and effective methods to address this challenge. This technique can be formulated as a form of Min-Max Optimization \cite{madry2017towards}, where the model parameters are iteratively updated to minimize training loss while simultaneously dealing with adversarial perturbations created by optimizing a specific adversary loss function. In recent years, there has been considerable focus on exploring how adversarial training affects the robustness of various self-supervised learning approaches \cite{chen2020adversarial, ghofrani2023rethinking}. Kim et al. \cite{kim2020adversarial} pioneered the use of contrastive loss to create adversarial examples without relying on any labels, to strengthen the robustness of the SimCLR \cite{chen2020simple}. Subsequently, Moshavash et al. \cite{moshavash2021momentum}, Wahed et al. \cite{wahed2022adversarial}, and Gowal et al. \cite{gowal2021self} adopted this approach for Momentum Contrast \cite{he2020momentum}, SwAV \cite{caron2020unsupervised}, and BYOL \cite{grill2020bootstrap}, respectively. Fan et al. \cite{fan2021does} introduced an additional regularization term within the contrastive loss to improve the transferability of cross-task robustness. They employed a method of generating pseudo-labels during adversarial training for downstream tasks. Similarly, Jiang et al. \cite{jiang2020robust} explored the robustness under different pair selection scenarios by considering a linear combination of two contrastive loss functions. 
\begin{table*}[t]
\centering
\small
\caption{\textbf{CF-AMC-SSL trains efficiently in fewer epochs, thereby reducing overall training time. By effectively employing multi-crop augmentations during base encoder training, it enhances both clean accuracy and robustness against PGD attacks.} Note that the highest values are indicated in red, while the second highest values are highlighted in
blue.}
\label{table: Comparison}
\renewcommand{\arraystretch}{1.3}
\resizebox{0.95\textwidth}{!}{%
\begin{tabular}{c*{6}{c}cc}
\toprule
\multicolumn{1}{c}{\textbf{Models}} & \multicolumn{3}{c}{\textbf{CIFAR-10}} & \multicolumn{3}{c}{\textbf{CIFAR-100}} & \multicolumn{2}{c}{\textbf{Time}} \\
\cmidrule(r){1-1} \cmidrule(lr){2-4} \cmidrule(lr){5-7} 
\textbf{Base Encoder} & \textbf{Clean} & \textbf{PGD(4/255)} & \textbf{PGD(8/255)} & \textbf{Clean} & \textbf{PGD(4/255)} & \textbf{PGD(8/255)} & \textbf{(min)} \\
\midrule
\midrule
\makecell{Patch-based EMP-SSL (baseline)\\{\tiny (16 patches, 5-step PGD, 30 epochs)}}&61&37.65 &16.95& 39.26&14.38 &4.22& 530 \\
 \makecell{Crop-based EMP-SSL\\{\tiny (16 crops,  5-step PGD, 30 epochs)}} &\textcolor{red}{76.55}   &\textcolor{blue}{53.3}  & \textcolor{blue}{28.49} &\textcolor{red}{51.71} &\textcolor{red}{33.88}  &\textcolor{red}{19.35} & 530 \\
\makecell{Crop-based SimCLR (baseline)\\{\tiny (2 crops, 5-step PGD, 500 epochs)}} &72.86 &47.98  &16.81&44.57  &19.84  &5.68 & 934 \\
 \makecell{Patch-based SimCLR\\{\tiny (2 patches, 5-step PGD, 500 epochs)}}&65.44&41.85  &17.19 &43.71 &21.87  &8.33  &934   \\
 \makecell{Patch-based EMP-FreeAdv\\{\tiny (16 patches, m=3, 10 epochs)}}&61.83 & 42.28& 21.53& 40.31 &23.78 & 12.13& \textcolor{red}{97} \\
 \makecell{Crop-based SimCLR-FreeAdv\\{\tiny (2 crops, m=3, 167 epochs)}} &70.25 &48.34 &24.5 &47.64 &26.53 &11.7& \textcolor{blue}{157}\\
\makecell{Crop-based EMP-FreeAdv (CF-AMC-SSL)\\{\tiny (16 crops, m=3, 10 epochs)}} &\textcolor{blue}{75.88}&\textcolor{red}{55.97}& \textcolor{red}{33.34}&\textcolor{blue}{50.74} & \textcolor{blue}{31.73}  &\textcolor{blue}{17.19} &\textcolor{red}{97} \\
\bottomrule
\end{tabular}
}
\end{table*}

Despite advancements in self-supervised learning (SSL), achieving adversarial robustness remains challenging due to two primary factors:  
1. \textbf{High computational cost:} Adversarial training requires additional gradient computations to generate adversarial examples, significantly increasing training time.  
2. \textbf{Trade-off between clean accuracy and robustness: }Adversarial training often compromises clean accuracy to improve robustness.

Extreme-Multi-Patch Self-Supervised Learning (EMP-SSL)~\cite{tong2023emp} offers a promising approach by reducing training epochs and leveraging fixed-size image patches instead of traditional multi-scale cropping. Building on EMP-SSL, this work explores the interplay between adversarial training and SSL, focusing on the following central questions:

\begin{enumerate}
    \item \textbf{Can multiple crops or patches compensate for fewer training epochs in self-supervised adversarial training?}  
    We examine whether increasing data diversity via multiple crops or patches can reduce computational overhead while maintaining performance.

    \item \textbf{How does crop diversity impact clean accuracy and robustness?}  
    By integrating EMP-SSL’s mechanisms with adversarial training, we analyze whether multiple crops can better balance the trade-off compared to SimCLR, which uses only two augmentations per image.

    \item \textbf{What is the impact of augmentation strategy (multi-scale crops vs. fixed-size patches) on robustness?}  
    We investigate how adopting diverse cropping strategies affects model robustness within the EMP-SSL framework.

     \item \textbf{How do different evaluation strategies perform for adversarially trained models?}  
    We compare robust Multi-Crop Embedding Aggregation (averaging embeddings from multiple crops) with the standard linear classifier applied to a single whole-image embedding.
    \item \textbf{Can free adversarial training improve adversarial SSL?}  
    We evaluate whether free adversarial training~\cite{shafahi2019adversarial}, known for its efficiency in supervised learning, can achieve competitive robustness in SSL under reduced training epochs.
\end{enumerate}

We selected SimCLR as a baseline because it is a standard SSL method that uses two augmentations per image and requires hundreds of epochs to converge. Its simplicity and ubiquity make it an ideal benchmark for contrasting the efficiency of EMP-SSL, which leverages augmentation diversity and fewer training epochs. By systematically comparing these two contrasting methods, our study explores how augmentation scaling and training efficiency impact adversarial robustness.

\textbf{Key Findings:}
\begin{itemize}
    \item Increasing the number of multi-scale crops effectively offsets fewer training epochs, enabling faster training without compromising performance. Although each epoch may require more time due to the higher number of crops or patches, the overall training time can be significantly reduced by decreasing the number of epochs. This approach leverages enhanced data augmentation to improve model efficiency while maintaining performance (Table \ref{table: Comparison} row 2 versus rows 3 and 4).
    \item Robust EMP-SSL with multi-scale crops achieves a better balance between clean accuracy and robustness than standard contrastive methods like SimCLR (Table \ref{table: Comparison} row 2 versus rows 3).
    \item Multi-scale crops within the robust EMP-SSL framework demonstrate notably superior outcomes for adversarial self-supervised learning when compared to fixed-size image patches \footnote{Note that robust central crop evaluation is likely to be less effective in terms of accuracy with fixed-scale patch-based pretraining because the model lacks exposure to the entire image during pretraining. On the other hand, robust multi-patch evaluation is time-intensive, as it necessitates generating multiple adversarial examples per image for the adversarial training of the linear classifier.} (Table \ref{table: Comparison} row 2 versus row 1).
    \item Central cropping outperforms robust Multi-Crop Embedding Aggregation in overall training time, clean accuracy, and robustness.
    \item Free adversarial training~\cite{shafahi2019adversarial} extends effectively to SSL, providing a cost-efficient solution even under reduced epochs. Our motivation for applying it arises from the observation that adversarially trained EMP-SSL already reduces the required training epochs. By integrating free adversarial training, we aim to further enhance efficiency, enabling us to examine its behavior with very few epochs (Table \ref{table: Comparison} row 7). Additionally, we explored its impact on SimCLR (see Table \ref{table: Comparison}, rows 5 and 6), offering a comprehensive analysis of its effects in SSL settings.
\end{itemize}

By systematically addressing these questions, our work advances adversarial SSL, offering efficient and robust algorithms tailored for real-world applications.

% These findings lead us to pose the following challenge.

% \textit{Can we achieve faster convergence in \textbf{robust} self-supervised learning, possibly within a few epochs, while simultaneously maintaining both robustness and clean accuracy?}"

% To address this question, we introduce an effective adversarial self-supervised learning method capable of converging in fewer than 10 epochs. We apply free adversarial training to the Crop-based EMP-SSL algorithm and refer to it as Cost-Free Adversarial Multi-Crop Self-Supervised Learning (CF-AMC-SSL). Through this approach, we demonstrate a significant reduction, by approximately two orders of magnitude, in the epochs required for adversarial SSL training.
% Our experiments solidly demonstrate the effectiveness of CF-AMC-SSL in improving both the clean accuracy and adversarial robustness (see Table \ref{table: Comparison}). For instance, CF-AMC-SSL achieves 75.88\% clean accuracy and defends against 20-step PGD attacks with 55.97\% and 33.34\% accuracy using ResNet-18, 4/255 and 8/255 $\infty$-norm perturbation strength, requiring just 97 minutes for adversarial training. In comparison, the SimCLR benchmark achieves 72. 86\% clean accuracy and defends with 47.98\% and 16.81\% accuracy under the same PGD attack settings on the CIFAR-10 data set, but it takes 934 minutes for adversarial training. In particular, CF-AMC-SSL improves both clean accuracy and adversarial robustness, accomplishing this with more than two orders of magnitude fewer training epochs.
\begin{figure}[t]
    \centering
    \begin{subfigure}{\linewidth}
        \centering
        \includegraphics[width=\linewidth ]{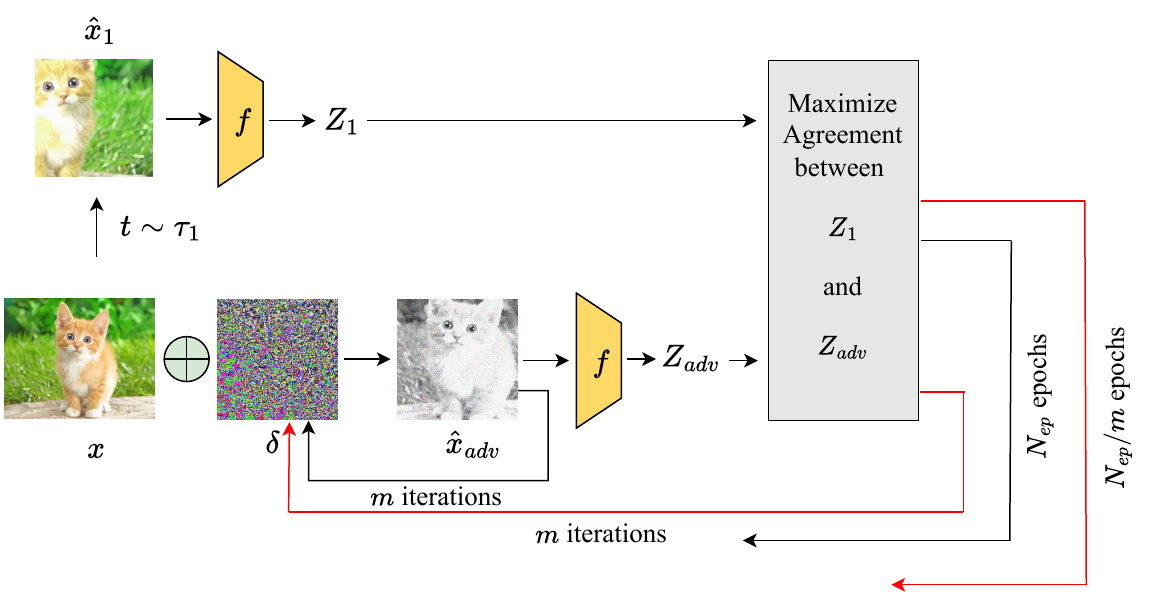}
        \caption{The adversarially trained SimCLR vs. free adversarially trained SimCLR framework.}
        \label{fig:contrastive}
    \end{subfigure}

    \hfill
    
    \begin{subfigure}{\linewidth}
        \centering
        \includegraphics[width=\linewidth]{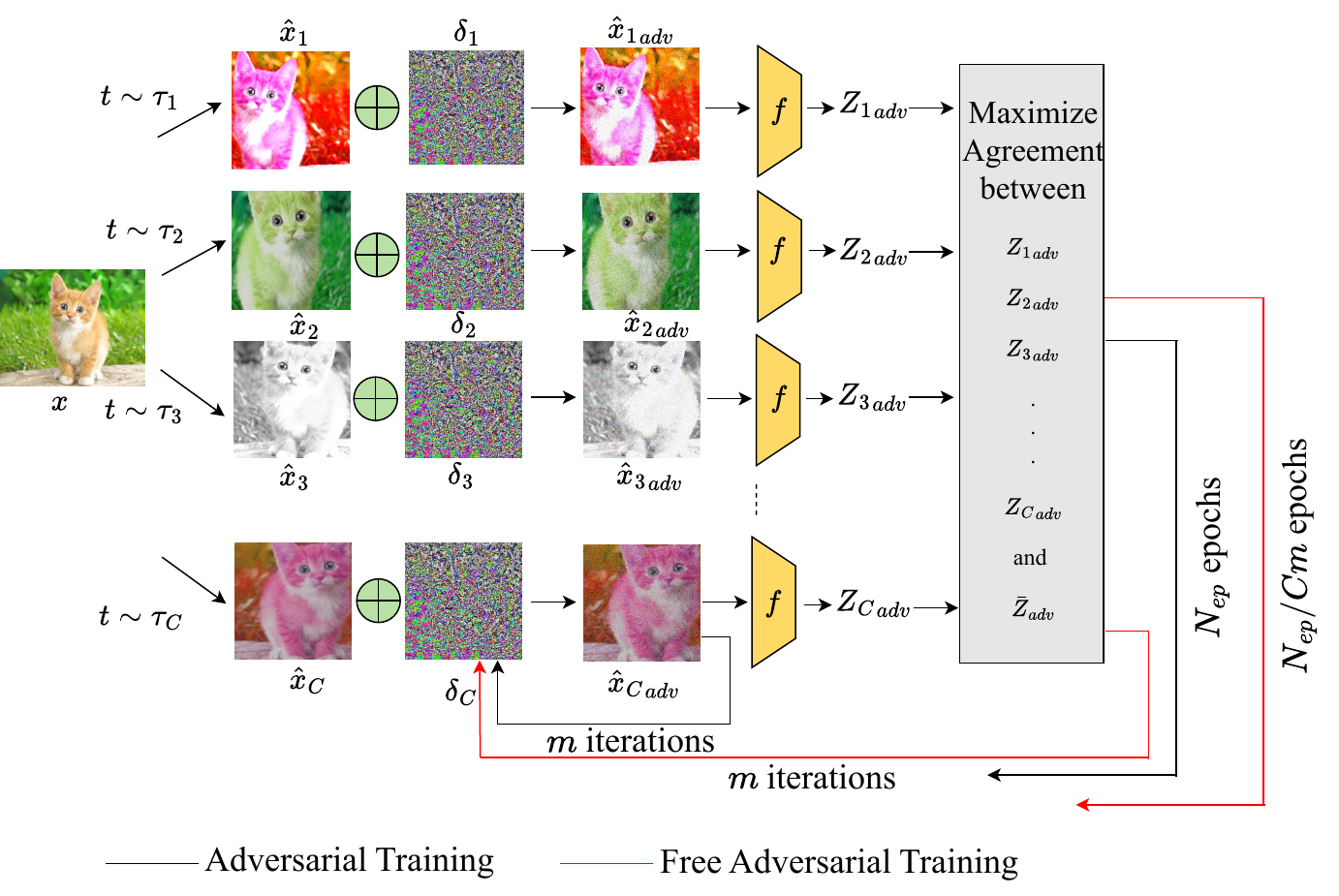}
        \caption{The adversarially trained crop-based EMP-SSL framework vs. the free adversarially trained crop-based EMP-SSL (CF-AMC-SSL).}
        \label{fig:EMP-SSL}
    \end{subfigure}

    \hfill
    
    \caption{Illustration of workflow comparison}

    \vspace{15 pt}
    
    \label{EMP}
\end{figure}

\section{Methodology}
As mentioned before, self-Supervised Learning (SSL) methods are designed to learn meaningful representations without relying on labeled data. A key challenge in SSL is preventing representation collapse, where embeddings become uniform and fail to capture the diversity of input features. To mitigate this, SSL methods typically employ either contrastive or information maximization strategies. Contrastive approaches, such as SimCLR\cite{chen2020simple} and MoCo \cite{he2020momentum}, contrast similar and dissimilar samples to encourage diverse representations. In contrast, non-contrastive methods, including VICReg\cite{bardes2021vicreg}, Barlow Twins\cite{zbontar2021barlow}, and TCR\cite{li2022neural}, leverage covariance regularization to maintain representation diversity without explicit negative sampling. Additionally, SwAV \cite{caron2020unsupervised}introduces multi-crop augmentations, combining views of varying resolutions to enhance performance, rather than relying solely on paired full-resolution views.

In this study, we focus on evaluating the robustness of Extreme-Multi-Patch Self-Supervised-Learning (EMP-SSL) \cite{tong2023emp}, a method that employs multiple augmentations per image and achieves convergence in significantly fewer epochs compared to standard SSL methods. We selected SimCLR as a baseline for its simplicity and its standard SSL design, which uses two augmentations per image and requires hundreds of epochs for convergence. This makes SimCLR a suitable reference point to assess the efficiency and robustness of EMP-SSL.

% In this section, we explain the methodology of our comparative study on the robustness of two learning schemes including baseline self-supervised contrastive learning (SimCLR) \cite{chen2020simple} and Extreme-Multi-Patch Self-Supervised-Learning (EMP-SSL) \cite{tong2023emp} algorithms.

\subsection{Background on SimCLR and EMP-SSL}

\subsubsection{SimCLR: A Simple Framework for Contrastive Learning}

SimCLR is a self-supervised learning framework designed to learn image representations by maximizing agreement between augmented views of the same image.

\subsubsection*{Key Components:}

\begin{itemize}
    \item \textbf{Data Augmentation:}  
    Positive pairs are generated through random augmentations such as cropping, color jittering, and Gaussian blur. Each input image \( x \) produces two augmented views, \( \hat{x}_1 \) and \( \hat{x}_2 \).  

    \item \textbf{Feature Extraction:}  
    A backbone network \( f(\cdot) \), typically a ResNet, maps each augmented view \( \hat{x}_i \) to a feature vector \( h_i \) in a high-dimensional space.  

    \item \textbf{Projection Head:}  
    A multi-layer perceptron (MLP) \( g(\cdot) \) projects the feature vector \( h_i \) into a lower-dimensional embedding space \( Z_i \), where the contrastive loss is applied:  
    \[
    Z_i = g(f(\hat{x}_i)).
    \]  

    \item \textbf{Contrastive Loss:}  
    The loss function encourages embeddings of positive pairs to be similar while pushing apart embeddings of negative pairs. The contrastive loss is defined as:  
    \[
    \mathcal{L}_{\text{contrastive}} = - \log \frac{\exp(\text{sim}(Z_1, Z_2)/\tau)}{\sum_{k=1}^{2N} \exp(\text{sim}(Z_1, Z_k)/\tau)},
    \]  
    where:
    \begin{itemize}
        \item \( \text{sim}(Z_1, Z_2) \) is the cosine similarity between embeddings \( Z_1 \) and \( Z_2 \),
        \item \( \tau \) is a temperature parameter, and
        \item \( N \) is the number of images in a mini-batch.
    \end{itemize}
\end{itemize}

SimCLR's framework efficiently captures meaningful representations by leveraging this contrastive learning approach.

\subsubsection{EMP-SSL: Extreme-Multi-Patch Self-Supervised Learning}

EMP-SSL extends the self-supervised learning paradigm by incorporating multiple fixed-scale patches per image and introducing novel loss terms to enhance representation consistency and generalization.

\subsubsection*{Key Components:}

\begin{itemize}
    \item \textbf{Multi-Patch Representation Learning:}  
    Each input image \( x \) is divided into \( C \) fixed-scale patches \( \{\hat{x}_1, \hat{x}_2, \dots, \hat{x}_C\} \). These patches are independently augmented and passed through a shared encoder \( h(\cdot) \) and a projection head \( g(\cdot) \) to produce embeddings:  
    \[
    Z_i = g(h(\hat{x}_i)),
    \]  
    where \( Z_i \) is the embedding of the \( i \)-th patch.

    \item \textbf{Invariance Term:}  
    To encourage consistency among embeddings, an invariance term \( D(Z_i, \bar{Z}) \) aligns each patch's embedding \( Z_i \) with the average embedding \( \bar{Z} \):  
    \[
    \bar{Z} = \frac{1}{C} \sum_{i=1}^C Z_i, \quad D(Z_i, \bar{Z}) = \text{Tr}((Z_i)^T \bar{Z}).
    \]  

    \item \textbf{Regularization Term:}  
    A regularization term \( R(Z_i) \) penalizes redundancy in feature embeddings by reducing correlations among dimensions:  
    \[
    R(Z_i) = \frac{1}{2} \log \det \left( I + \frac{d}{b\epsilon^2} Z_i Z_i^T \right),
    \]  
    where \( Z_i Z_i^T \) is the covariance matrix of the embeddings.

    \item \textbf{Overall Objective Function:}  
    EMP-SSL optimizes the sum of the invariance and regularization terms across all patches:  
    \[
    \mathcal{L}_{\text{EMP-SSL}} = \sum_{i=1}^C \left[ D(Z_i, \bar{Z}) + R(Z_i) \right].
    \]  
\end{itemize}

By leveraging these innovative loss terms, EMP-SSL fosters stronger invariance and diversity in learned representations, making it a powerful approach for self-supervised learning.

\subsection{Extending SimCLR and EMP-SSL with Adversarial Training}
Initially, our analysis in Table \ref{table: ST-simclr-vs-EMP} reveals that both the baseline EMP-SSL and SimCLR \textbf{without adversarial training} are vulnerable to adversarial attacks. This finding further underscores the lack of robustness in SimCLR, consistent with previous research findings \cite{ghofrani2023rethinking, kim2020adversarial}. 

We extend the SimCLR and EMP-SSL frameworks by incorporating adversarial training to improve robustness against adversarial attacks, as depicted in Figure \ref{EMP}. Adversarial training integrates adversarial examples into the learning process, enabling models to better withstand perturbations. Below, we describe the generation of adversarial examples and the modifications to the training objectives for both frameworks.

\subsubsection{Robust SimCLR: Adversarial Contrastive Learning}

In the robust version of SimCLR, adversarial training enhances the model's resilience. For each image in the mini-batch, adversarial examples are generated using PGD attacks. Both the original augmented view and its adversarial counterpart are treated as positive pairs. The contrastive loss function is updated to maintain similarity between these pairs while distinguishing them from negative samples, thereby reinforcing the model’s ability to generalize under adversarial perturbations.

\subsubsection{Robust EMP-SSL: Adversarial Multi-Patch Learning}

In robust EMP-SSL, adversarial training is integrated to strengthen the model's defenses. The representation for each image is aggregated from multiple crops or patches, and adversarial perturbations are applied independently to each crop/patch.

\paragraph{\textbf{Adversarial Perturbation Process:}}
\begin{enumerate}
    \item The image is divided into multiple crops or patches.
    \item An adversarial perturbation is generated and updated independently for each crop or patch.
    \item Each crop is perturbed individually, rather than applying a shared perturbation across all crops.
\end{enumerate}

This independent generation of adversarial examples results in more diverse and challenging perturbations, boosting the model’s robustness.

\paragraph{\textbf{Updated Training Objective:}}
The training objective includes:
\begin{itemize}
    \item \textbf{Regularization Term} $R(Z_{\text{adv}, i})$: Penalizes high correlations among adversarial representations.
    \item \textbf{Consistency Term} $D(Z_{\text{adv}, i}, \bar{Z}_{\text{adv}})$: Promotes consistency among adversarial embeddings.
\end{itemize}

These terms regularize adversarial representations while maintaining their similarity to the original augmented views, preventing overfitting to adversarial examples.

\subsubsection{Adversarial Training Strategies: Crop-Based vs. Patch-Based Methods}

We evaluate two augmentation strategies for adversarial training in robust EMP-SSL:
\begin{enumerate}
\item \textbf{Crop-Based Method:} Random crops are taken from the augmented image, with crop sizes ranging from $9 \times 9$ to $32 \times 32$ pixels. This approach leverages the diversity of random crops, enhancing robustness by forcing the model to generalize across a wide range of image regions.
\item \textbf{Patch-Based Method:} Fixed-scale patches are extracted at predefined scales from the image. While simpler, this method may not achieve the same level of robustness as the crop-based approach due to limited variability.
\end{enumerate}

\begin{table*}[t]
\centering
\small
\caption{\textbf{Comparative results of clean data performance and robustness against PGD attacks: baseline SimCLR versus EMP-SSL with standard pretraining on CIFAR10 and CIFAR100 datasets.}}
\label{table: ST-simclr-vs-EMP}
\renewcommand{\arraystretch}{1.3}
\resizebox{\textwidth}{!}{%
\begin{tabular}{cc*{16}{c}}
\toprule
\multicolumn{2}{c}{\textbf{Models}} & \multicolumn{4}{c}{\textbf{CIFAR-10}} & \multicolumn{4}{c}{\textbf{CIFAR-100}} \\
\cmidrule(r){1-2} \cmidrule(lr){3-6} \cmidrule(l){7-10} \cmidrule(l){11-14}
\textbf{Linear Classifier} & \textbf{Base Encoder} & \textbf{Clean} & \textbf{PGD(4/255)} & \textbf{PGD(8/255)} & \textbf{PGD(16/255)} & \textbf{Clean} & \textbf{PGD(4/255)} & \textbf{PGD(8/255)} & \textbf{PGD(16/255)} \\
\midrule
\midrule
\multirow{2}{*}{Central Crop} & SimCLR & 86.65 & 0.13 & 0 & 0 & 62.5 & 0.74 & 0.53 & 0.45 \\
 & EMP-SSL & 75.02 & 0 & 0 & 0 & 44.31 & 0.02 & 0.02 & 0.02 \\
 \midrule
% \multirow{2}{*}{1 Crop (Patch)} & SimCLR &  70.9 & 0.07 & 0.01 & 0 & 43.42 & 0.06 & 0.01 & 0.01 \\
%  & EMP-SSL &  80.73 & 0.41 & 0.29 & 0.24& 53.62 & 0.45 & 0.22 & 0.12 \\
%  \midrule
\multirow{2}{*}{32 Crops (Patchs)} & SimCLR & 86.68 & 0.02 & 0 & 0 & 65.21 & 0.18 & 0.12 & 0.07 \\
 & EMP-SSL & 92.85 & 0.04 & 0.01 & 0 &  71.82 & 0.46 & 0.15 & 0.08 \\
 \midrule
\multirow{2}{*}{64 Crops (Patches)} & SimCLR & 89.31 & 0.01 & 0 & 0 & 66.3 & 0.17 & 0.12 & 0.12 \\
 & EMP-SSL & 93.29 & 0.02 & 0.03 & 0.01 & 72.3 & 0.5 & 0.2 & 0.09 \\
\bottomrule
\end{tabular}
}
\end{table*}

\begin{figure*}[htbp]
    \centering
    \includegraphics[width=\textwidth]{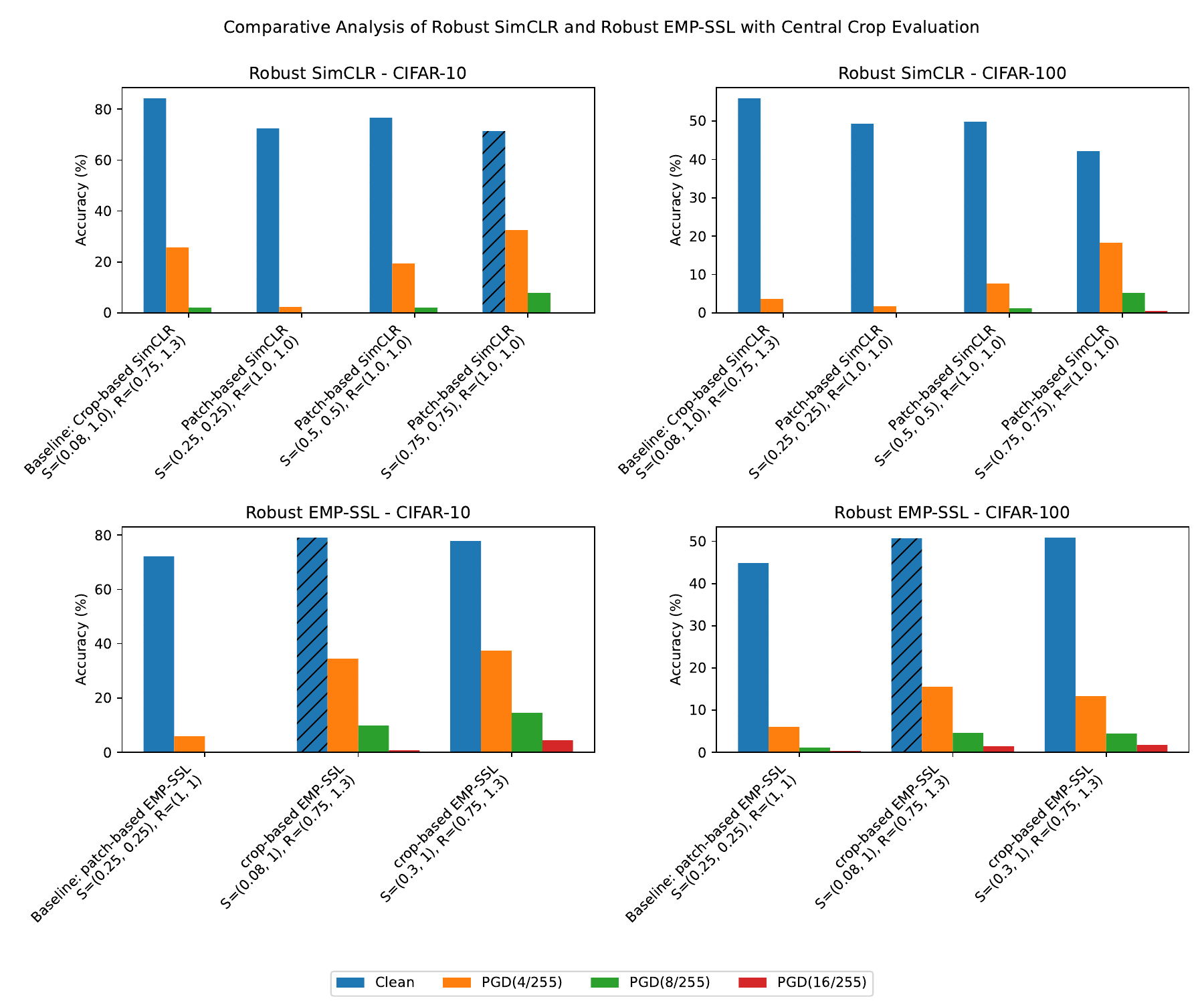}  % Replace with your PDF file name
   \caption{\textbf{Evaluation of robustness against PGD attacks through adversarial pretraining on CIFAR-10 and CIFAR-100 datasets.} We compare the performance of robust SimCLR and robust EMP-SSL with central crop evaluation under different training configurations. Our analysis includes the evaluation of patch-based SimCLR with varying patch sizes and baseline SimCLR, revealing a noticeable trade-off between clean accuracy and robustness. Larger patch sizes in robust SimCLR improve robustness but reduce clean accuracy. Additionally, we compare crop-based EMP-SSL (with varying crop sizes) to baseline EMP-SSL, demonstrating that the crop-based approach significantly enhances robustness. Notably, Robust EMP-SSL achieves a superior balance between clean accuracy and robustness compared to robust SimCLR. The variables $S$ and $R$ correspond to the scales and ratios used in the PyTorch framework’s RandomResizedCrop method.}
    \label{eval-simclr-emp-central}
\end{figure*}

\section{Experiments and Results}
As mentioned above, EMP-SSL significantly reduces training epochs in self-supervised learning by increasing the number of fixed image patches augmented. This approach diverges from traditional methods like SimCLR, which rely on extensive epochs and utilize only two augmented multi-scale image crops per image. Through a comprehensive analysis of EMP-SSL within an adversarial training framework, we seek insights into the relationship between training epochs, image crop choices, model robustness, and accuracy.
Our focus is directed towards presenting the results through linear probing accuracy. This evaluation comprises two essential components: 
\begin{enumerate}
\item \textbf{Standard Central Crop Assessment:} This conventional method trains and evaluates a linear classifier using a single fixed central patch from each image, where the entire image serves as the central patch.
\item \textbf{Multi-Patch (Multi-Crop) Embedding Aggregation Evaluation:} In contrast to the standard central crop assessment, an alternative method involves creating an image embedding by inputting a specified number of crops (patches) into the base encoder. These crop (patch) embeddings are then combined through averaging and fed into the linear classifier. Henceforth, we will refer to this evaluation method as the \textbf{"$\bm{n}$ Crops (Patches)"} linear classifier. Note that the patches (crops) are sampled with the same scale factor as during the pretraining phase. 
\end{enumerate}
These evaluation methods provide comprehensive insights into the model's capabilities, considering both accuracy and robustness, utilizing conventional practices and innovative multi-patch and multi-crop embedding strategies.
To evaluate the model's robustness, we employ a threat model in which the adversary possesses full knowledge of the base encoder's architecture and network parameters as well as those of the linear classifier. The attacks are generated end-to-end by leveraging the cross-entropy loss function.

\subsection{Experiment Setup}\label{Sec:results}
Our experimental setup is similar to that used in \cite{tong2023emp}. We employ a ResNet-18 as the base encoder for all experiments unless otherwise specified. We typically train models for 30 epochs when using EMP-SSL and for 500 epochs when utilizing SimCLR as EMP-SSL converges more quickly, unless explicitly stated otherwise.
In all adversarial training scenarios, adversarial perturbations are generated using a 5-step PGD attack under the $l_\infty$ norm with a maximum perturbation limit of $\epsilon = 8/255$, unless a specific value of $\epsilon$ is specified. The models are evaluated against 20-step PGD attacks. Furthermore, we include results against Auto-Attack \cite{croce2020reliable} at the end to confirm our findings. We report the top-1 test accuracy for all scenarios to evaluate their robustness. After sampling the patches and crops, we ensure that their resolution matches the original image sizes before feeding them into ResNet-18 for embedding. It is important to note that our objective is not to evaluate transfer learning capabilities (cross-dataset validation). Consequently, in all experiments, both the base encoder and the linear classifier are trained on the same dataset.

\subsection{Both SimCLR and EMP-SSL are vulnerable to adversarial attacks through standard training}

We start by performing standard pretraining on the base encoder, following the methodology of SimCLR and EMP-SSL. This experiment aims to achieve two objectives: 1- to investigate whether employing multiple fixed-sized patches for training the base encoder in SSL, like EMP-SSL, can enhance its robustness against adversarial examples, without resorting to adversarial training, as compared to methods like SimCLR, which utilize only one pair of differently scaled crops. 2- to evaluate the robustness of the standard linear classifier (central crop) versus the evaluation based on $n$ patches (crops).

The detailed results of these evaluations, including the performance on clean data and the robustness against PGD attacks, are presented in Table \ref{table: ST-simclr-vs-EMP} for both the CIFAR10 and CIFAR100 datasets. During the evaluation phase, we present results for standard central crop evaluation in both crop-based and patch-based methods. Additionally, we employ multi-patch (multi-crop) aggregated evaluation. Here, the term `patch' signifies the aggregation of fixed-scale patch embeddings, while `crop' indicates the aggregation of multi-scale crop embeddings. Importantly, when training the base encoder with crops (patches), we apply the same crops (patches) consistently during both the training and testing phases of the linear classifier. The results highlight an interesting observation:

\textit{While the combination of multi-patch aggregated evaluation with EMP-SSL can indeed enhance model performance on clean data, it is evident that this learning method remains vulnerable to adversarial examples, similar to the behavior observed in the SimCLR method.}

\begin{table*}[htb]
\centering
\small
\caption{\textbf{Evaluation of Robust EMP-SSL and Robust SimCLR across different adversarial training scenarios on CIFAR-10 and CIFAR-100 datasets:} The findings imply that boosting the robustness of the linear classifier contributes to enhancing the overall robustness of both the robust SimCLR and EMP-SSL.}
\label{table:r-LE}
\renewcommand{\arraystretch}{1.3}
\resizebox{\textwidth}{!}{%
\begin{tabular}{cc*{16}{c}}
\toprule
\multicolumn{2}{c}{\textbf{Models}} & \multicolumn{4}{c}{\textbf{CIFAR-10}} & \multicolumn{4}{c}{\textbf{CIFAR-100}} \\
\cmidrule(r){1-2} \cmidrule(lr){3-6} \cmidrule(lr){7-10} \cmidrule(l){11-14}
\textbf{Linear Classifier} & \textbf{Base Encoder} & \textbf{Clean} & \textbf{PGD(4/255)} & \textbf{PGD(8/255)} & \textbf{PGD(16/255)} & \textbf{Clean} & \textbf{PGD(4/255)} & \textbf{PGD(8/255)} & \textbf{PGD(16/255)} \\
\midrule
\midrule
\multirow{2}{*}{Central Crop} & Robust SimCLR & \textbf{84.24} & 25.68 & 1.97 & 0.07 & \textbf{55.91} & 3.58 & 0.18 & 0 \\
 & \makecell{Robust EMP-SSL\\{\small Crop-based(16)}} & 80.72 &  33.62 &  8.95 &  0.92 &  51.83 &  19.3 &  6.85 &  1.73 \\
 \midrule
\multirow{2}{*}{Robust Central Crop (r-LE)} & Robust SimCLR &  72.86 &47.98  &16.81 &0.33&44.57  &19.84  &5.68&0.26 \\
 & \makecell{Robust EMP-SSL\\{\small Crop-based(16)}} &76.55 & \textbf{53.3} & \textbf{28.49} & \textbf{3.96} &  51.71 &  \textbf{33.88} & \textbf{19.35} & \textbf{4.92}\\
\bottomrule
\end{tabular}
}
\end{table*}

\sethlcolor{pink}
\begin{table*}[t]
\centering
\small
\caption{\textbf{Evaluation of CF-AMC-SSL and SimCLR-FreeAdv algorithms:} The results show that CF-AMC-SSL trains efficiently in fewer epochs, reducing the overall training time. Additionally, employing multi-crop augmentations in CF-AMC-SSL during base encoder training effectively improves both accuracy and robustness. Note that the topmost and the second-highest values are indicated in red and blue, respectively.}
\label{table:CF-AMC-SSL}
\renewcommand{\arraystretch}{1.3}
\resizebox{\textwidth}{!}{%
\begin{tabular}{cc*{8}{c}c|c*{4}{c}}
\toprule
\multicolumn{2}{c}{\textbf{Models}} & \multicolumn{4}{c}{\textbf{CIFAR-10}} & \multicolumn{4}{c}{\textbf{CIFAR-100}} & \textbf{Time} & \multicolumn{4}{|c}{\textbf{ImageNet-100}} \\
\cmidrule(r){1-2} \cmidrule(lr){3-6} \cmidrule(lr){7-10} \cmidrule(l){12-15} 
\textbf{\makecell{Linear\\ Classifier}} & \textbf{\makecell{Base Encoder\\ResNet-18}} & \textbf{Clean} & \textbf{\makecell{PGD\\4/255}} & \textbf{\makecell{PGD\\8/255}} & \textbf{\makecell{PGD\\16/255}} & \textbf{Clean} & \textbf{\makecell{PGD\\4/255}} & \textbf{\makecell{PGD\\8/255}} & \textbf{\makecell{PGD\\16/255}} & \textbf{(min)} & \textbf{Clean} & \textbf{\makecell{PGD\\4/255}} & \textbf{\makecell{PGD\\8/255}} & \textbf{\makecell{PGD\\16/255}} \\
\midrule
\midrule
\multirow{11}{*}{\makecell{Robust\\Central\\Crop}} 
 & \makecell{Crop-based EMP-SSL\\{\tiny (16 crops, 5-step PGD, 30 epochs)}}& \textcolor{red}{76.55} & 53.3 & 28.49 & 3.96 & \textcolor{red}{51.71} & \textcolor{red}{33.88} & \textcolor{red}{19.35} & \textcolor{blue}{4.92} & 530 & -- & -- & -- & -- \\
 & \makecell{CF-AMC-SSL\\{\tiny (16 crops, m=3, 10 epochs)}} &\textcolor{blue}{75.78} & \textcolor{red}{55.97} & \textcolor{blue}{33.34}&\textcolor{blue}{6.24} & \textcolor{blue}{50.74} & \textcolor{blue}{31.73} & 17.19 & 3.49 & \textcolor{red}{97} & -- & -- & -- & -- \\
  & \makecell{CF-AMC-SSL\\{\tiny (16 crops, m=5, 6 epochs)}} & 71.89 & \textcolor{blue}{54.2} & \textcolor{red}{34.94} & \textcolor{red}{8.55} & 45.84 & 30.1 & \textcolor{blue}{17.84} & \textcolor{red}{5.15} & \textcolor{red}{97} & 34.38 & 18.82 & 8.22 & 1.2 \\
   & \makecell{CF-AMC-SSL\\{\tiny (16 crops, m=5, 10 epochs)}} & - & - & - & - & - & - & - & - & - & 46.26 & 27.86 & 13.94 & 2.16 \\
 & \makecell{Crop-based SimCLR\\{\tiny (5-step PGD, 500 epochs)}}& 72.86& 47.98 & 16.81 & 3.3 & 44.58 & 19.84 & 5.68 & 0.26 & 934 & -- & -- & -- & -- \\
 & \makecell{SimCLR-FreeAdv\\{\tiny (m=3, 167 epochs)}} &70.25 & 48.34 & 24.5 & 2 & 47.64 & 26.53 & 11.7& 1.26 & 157 & 30.26 & 16.14 & 6.06 & 0.25 \\
  & \makecell{SimCLR-FreeAdv\\{\tiny (m=5, 100 epochs)}} &69.97 & 51.36 & 30.84 & 5.7& 45.69 & 29.43 & 16.15 & 3.1 & 157 & 25.26 & 14.12 & 5.88 & 0.7 \\
  \cmidrule(lr){2-15}
  & \makecell{CF-AMC-SSL\\{\tiny (16 crops, m=5, \textbf{18 epochs})}} & 76.28 & 58.06 & 37.5 & 9.39& 52.01 & 33.3 & 19.34 & 5.06 & 291 & -- & -- & -- & -- \\
  & \makecell{CF-AMC-SSL\\{\tiny (16 crops, \textbf{m=12}, 10 epochs)}} & 55.84 & 43 & 30.84 & \textbf{12.4} & 31.33 & 21.86 & 14.62 & \textbf{6.14} & 388 & -- & -- & -- & -- \\
  
\midrule
\multicolumn{2}{c}{\makecell{Supervised-FreeAdv\\{\tiny (m=3, 300 epochs)}}} & 82.63 & 47.12 & 16.27& 1.3 & 52.07 & 20.2 & 6.34 & 0.92 & 155 & -- & -- & -- & -- \\
\multicolumn{2}{c}{\makecell{Supervised-FreeAdv\\{\tiny (m=7, 300 epochs)}}} & 74.63 & 48.56 & 23.75& 2.87 & 39.88 & 19.97 & 8.14 & 1.12 & 360 & -- & -- & -- & -- \\
\bottomrule
\end{tabular}
}
\end{table*}

\begin{table*}[t]
\centering
\small
\caption{\textbf{Evaluation of Different Learning Algorithms Using ResNet-50 as the Base Encoder:}
This experiment demonstrates the generalizability of our findings when employing a larger base encoder, such as ResNet-50. Additionally, it highlights that increasing the number of iterations for PGD enhances the model's robustness against larger perturbations.}
\label{table:CF-AMC-SSL-resnet50}
\renewcommand{\arraystretch}{1.3}
\resizebox{\textwidth}{!}{%
\begin{tabular}{cc*{8}{c}}
\toprule
\multicolumn{2}{c}{\textbf{Models}} & \multicolumn{4}{c}{\textbf{CIFAR-10}} & \multicolumn{4}{c}{\textbf{CIFAR-100}} \\
\cmidrule(r){1-2} \cmidrule(lr){3-6} \cmidrule(lr){7-10}
\textbf{\makecell{Linear\\ Classifier}} & \textbf{\makecell{Base Encoder\\ResNet-50}} & \textbf{Clean} & \textbf{PGD(4/255)} & \textbf{PGD(8/255)} & \textbf{PGD(16/255)} & \textbf{Clean} & \textbf{PGD(4/255)} & \textbf{PGD(8/255)} & \textbf{PGD(16/255)} \\
\midrule
\midrule
\multirow{4}{*}{\makecell{Robust \\Central\\ Crop}} & \makecell{CF-AMC-SSL\\{\tiny (16 crops, m=7, 9 epochs)}} &73.43 & 56.81 & 38.31 & 11.03 & 47.4 & 32.19 & 19.9& 6.21 \\
  & \makecell{CF-AMC-SSL\\{\tiny (16 crops, m=3, 10 epochs)}} &75.89 & 57.61 &35.19 & 6.51 & 53.34 & 33.08 & 18.15 & 3.72 \\
 & \makecell{SimCLR-FreeAdv\\{\tiny (m=7, 150 epochs)}} &49.49 & 39.09 & 28.88 & 12.76 & 25.49 & 18.36 & 12.51 & 5.51 \\
  & \makecell{SimCLR-FreeAdv\\{\tiny (m=3, 150 epochs)}} &66.89& 47.52 & 26.9 & 3.68& 38.95& 24.14 & 12.17 & 1.61 \\
\bottomrule
\end{tabular}
}
\end{table*}

\begin{table*}[t]
\centering
\small
\caption{\textbf{Evaluation of different learning algorithms against AutoAttack (AA):} The Autoattack evaluation confirms that using multi-crop augmentations in CF-AMC-SSL during base encoder training improves both accuracy and robustness effectively. Note that the topmost and the second-highest values are indicated in red and blue, respectively.}
\label{table:CF-AMC-SSL-Autoattack}
\renewcommand{\arraystretch}{1.3}
\resizebox{\textwidth}{!}{%
\begin{tabular}{cc*{8}{c}}
\toprule
\multicolumn{2}{c}{\textbf{Models}} & \multicolumn{4}{c}{\textbf{CIFAR-10}} & \multicolumn{4}{c}{\textbf{CIFAR-100}} \\
\cmidrule(r){1-2} \cmidrule(lr){3-6} \cmidrule(lr){7-10}
\textbf{\makecell{Linear\\ Classifier}} & \textbf{\makecell{Base Encoder\\ResNet-18}} & \textbf{Clean} & \textbf{AA(4/255)} & \textbf{AA(8/255)} & \textbf{AA(16/255)} & \textbf{Clean} & \textbf{AA(4/255)} & \textbf{AA(8/255)} & \textbf{AA(16/255)} \\
\midrule
\midrule
\multirow{6}{*}{\makecell{Robust \\Central\\ Crop}} & \makecell{Crop-based EMP-SSL\\{\tiny (16 crops, 5-step PGD, 30 epochs)}}& \textcolor{red}{76.55} & 23.93 & 26.57 & 7.81 & \textcolor{red}{51.71} & \textcolor{red}{33.88} & \textcolor{red}{19.35} &4.92 \\
 & \makecell{CF-AMC-SSL\\{\tiny (16 crops, m=3, 10 epochs)}} &\textcolor{blue}{75.78} & \textcolor{red}{51.52} & \textcolor{blue}{ 27.74} & 6.59 & \textcolor{blue}{50.74} & \textcolor{blue}{27.05} & 15.35& 5.23 \\
  & \makecell{CF-AMC-SSL\\{\tiny (16 crops, m=5, 6 epochs)}} &71.89 & \textcolor{blue}{50.76} & \textcolor{red}{30.14} & \textcolor{blue}{8.44} & 45.84 & 26.99 & \textcolor{blue}{16.3} & \textcolor{red}{5.83} \\
 & \makecell{Crop-based SimCLR\\{\tiny (5-step PGD, 500 epochs)}}& 72.86 & 16.66 & 12.57 & \textcolor{red}{10.5} & 44.58 & 8.81 & 6.21 & \textcolor{blue}{5.29} \\
 & \makecell{SimCLR-FreeAdv\\{\tiny (m=3, 167 epochs)}} &70.25 & 46.62 & 22.31 & 4.76 & 47.64 & 23.35 & 13.6 & 3.93 \\
  & \makecell{SimCLR-FreeAdv\\{\tiny (m=5, 100 epochs)}} &69.97 & 48.91 & 27.51 & 6.32 & 45.69 & 25.39 & 14.28 & 5.12 \\
\bottomrule
\end{tabular}
}
\end{table*}

\subsection{Robust Crop-Based EMP-SSL Improves Both Clean Accuracy and Robustness Compared to Robust SimCLR}

In this experiment, we applied adversarial training to both SimCLR and EMP-SSL algorithms. For evaluation, we primarily focused on central cropping, which offers a balance between computational efficiency and accuracy. The crop-based and patch-based approaches differ in their augmentation strategies during training: the crop-based method employs multi-scale cropping, while the patch-based method utilizes fixed-scale patching augmentations. Additional evaluations using multi-patch aggregation with 32 and 64 patches, as well as detailed ablation studies, are provided in the Appendix.
Within the SimCLR framework, two random augmentations (crops or patches) per image are selected during training, while EMP-SSL generates 40 random patches (crops). Results for adversarially trained SimCLR and EMP-SSL under various training configurations are shown in Figure \ref{eval-simclr-emp-central}.

The key findings include:
\begin{itemize}
    \item \textit{Training base encoders with SimCLR using only two augmentations per image results in a notable trade-off between clean accuracy and robustness.}
    \item \textit{Crop-based EMP-SSL demonstrates greater robustness against adversarial attacks, whereas patch-based EMP-SSL excels on clean data}.
    \item \textit{Central cropping is computationally efficient and achieves strong clean accuracy and robustness, particularly when compared to crop (patch) embedding aggregation (detailed in the Appendix).}
    \item \textit{The ablation study (detailed in the Appendix) highlights that increasing the number of patches used during adversarial training improves clean accuracy in patch-based EMP-SSL. Additionally, a moderate number of crops (e.g., 16) in crop-based EMP-SSL maintains a better trade-off between clean accuracy and robustness under central cropping evaluation.}
    \item \textit{Robust EMP-SSL achieves a superior balance between clean accuracy and robustness when compared to robust SimCLR.}
\end{itemize}

\subsection{Robust Crop-Based EMP-SSL with Robust Linear Evaluation}
While the robust crop-based EMP-SSL demonstrates improved robustness in standard evaluation, we further scrutinize its robustness by employing an additional assessment known as robust linear evaluation (r-LE) \cite{kim2020adversarial}. This experiment is designed to evaluate the robustness of the model trained using the r-LE approach. This scenario involves training a base encoder using the robust crop-based EMP-SSL algorithm, followed by adversarial training of the linear classifier separately after freezing the base encoder. The results in Table \ref{table:r-LE} indicate that
\textit{\begin{itemize}
    \item Improving the robustness of the linear classifier contributes to improved model robustness for both the robust SimCLR and EMP-SSL algorithms.
    \item Robust crop-based EMP-SSL preserves a greater balance between clean accuracy and robustness of the model. 
\end{itemize}}

\begin{algorithm}[t]
  \SetAlgoLined
  \SetKwInOut{Initialize}{Initialize}
  \SetKwInOut{Iterate}{Iterate}
  \SetKwInOut{Perform}{Perform}
  \Initialize{$\delta \gets 0$}
  \textit{//Iterate $N_{ep}/m$ times to account for minibatch replays and run for $N_{ep}$ total epochs}{\\
    \For{$epoch = 1$ to $N_{ep}/m$}{
      \For{$i = 1$ to $D$}{
      \textit{//Augment data for CF-AMC-SSL learning:\\
        \For{$k = 1$ to $C$}{
            Draw augmentation function $t_k$\;
            $\hat{x}_{i,k} = t_k(x_i)$\;
            }
          \For{$j = 1$ to $m$}{
            \textit{//Compute gradients for perturbation and model weights simultaneously:\\
            $\nabla \delta, \nabla \theta = \nabla \mathcal{L_{EMP-SSL}}(f\circ g_{\theta}(\hat{x}_{i,k} + \delta))$\\
            $\delta = \delta + \epsilon \cdot \text{sign}(\nabla \delta)$ //Update $\delta$ with the gradients calculated\\
            $\delta = \max(\min(\delta, \epsilon), -\epsilon)$\\
            $\theta = \theta - \nabla \theta$ // Update model weights with some optimizer
          }
        }
      }
    }
  }
  }
  \caption{CF-AMC-SSL learning algorithm for $N_{ep}$ epochs, given some radius $\epsilon$, $m$ minibatch replays, $C$ number of crops, and a dataset of size $D$ for an encoder $f_\theta$ and a projector $g_\theta$}
  \label{algorithm}
\end{algorithm}

\subsection{Cost-Free Adversarial Multi-Crop Self-Supervised Learning Evaluation}
Inspired by these findings, we introduce an effective adversarial self-supervised learning method capable of converging in fewer than 10 epochs. We apply free adversarial training~\cite{shafahi2019adversarial} to the crop-based EMP-SSL framework (see Figure \ref{EMP}), referring to it as \textbf{Cost-Free Adversarial Multi-Crop Self-Supervised Learning (CF-AMC-SSL)}. This approach achieves a dramatic reduction---approximately two orders of magnitude---in the epochs required for adversarial SSL training (Algorithm \ref{algorithm}).

Our method employs an iterative approach for adversarial training of the base encoder, integrating multi-crop augmentations. Specifically, this strategy involves repeating each iteration $m$ times within a minibatch, reusing gradient information computed during parameter updates. This enables the generation of adversarial examples before progressing to the next iteration. Experimental results for various $m$ values are presented in Table \ref{table:CF-AMC-SSL}, alongside comparisons to the 5-step PGD adversarially trained crop-based EMP-SSL model, which demands approximately five times the training time of CF-AMC-SSL variants.

For a comprehensive evaluation, we incorporated free adversarial training into the SimCLR algorithm and its supervised variant (termed SimCLR-FreeAdv and Supervised-FreeAdv, respectively), with results summarized in Table \ref{table:CF-AMC-SSL}. All experiments were conducted on a single A6000 GPU, with runtime comparisons performed on CIFAR-10 and CIFAR-100 datasets. Although employing multi-crop augmentations in joint-embedding SSL might seem to increase training times---especially when generating adversarial examples---our findings highlight notable efficiency gains.

Robust EMP-SSL converges in significantly fewer epochs (30 epochs) compared to robust SimCLR (500 epochs) and requires less runtime (530 minutes vs. 934 minutes). Our CF-AMC-SSL framework advances this efficiency further, leveraging insights from free adversarial training \cite{shafahi2019adversarial} to reduce runtime to just 97 minutes---over five times faster than robust EMP-SSL---while achieving comparable performance. Furthermore, CF-AMC-SSL effectively balances clean accuracy and adversarial robustness, consistently outperforming both robust SimCLR and the robust supervised method, even when label information is available.

We extended our experiments to higher values of \(m\) and more training epochs, finding that increasing \(m\) improves robustness against stronger perturbations (e.g., \(\epsilon = 16/255\)). To evaluate generalizability, we also conducted experiments on the ImageNet-100 dataset. Despite its distinct characteristics, the results exhibited a similar trend, further confirming the robustness and efficiency of our approach. Additionally, we performed experiments using ResNet-50 as the base encoder to assess generalization (see Table \ref{table:CF-AMC-SSL-resnet50}). These results validate our approach, showing that using a larger network and more iterations for PGD attacks leads to enhanced robustness, especially for larger perturbations.

In summary, our framework offers two significant advantages:  
\begin{enumerate}
    \item \textbf{Efficient training:} Achieves model training with orders of magnitude fewer epochs, significantly reducing overall runtime.  
    \item \textbf{Enhanced accuracy and robustness:} Utilizes multi-crop augmentations during base encoder training to effectively improve both clean accuracy and adversarial robustness.
\end{enumerate}

Finally, we evaluated our models against Auto-Attack, as shown in Table \ref{table:CF-AMC-SSL-Autoattack}, further affirming the reliability of our findings.

\section{Discussion}

In this paper, we introduced CF-AMC-SSL, a crop-based EMP-SSL method enhanced with free adversarial training. This approach significantly improves the learning efficiency of robust self-supervised learning by leveraging the co-occurrence statistics of crops within images. We demonstrated that utilizing a moderate number of augmentations per image during training enhances adversarial robustness while maintaining clean accuracy, outperforming robust SimCLR, which uses only two augmentations per image. The increased number of image augmentations allows the model to encounter a broader range of perturbed images, altering style information (such as color, textures, and background) while preserving content. This process enables the model to extract representations more focused on content rather than style, thereby improving overall performance.

Similar to previous works \cite{chen2022intra,li2022neural}, our method defines the representation of a given image \( x \) as the average of the embeddings \( h_1, \ldots, h_n \) of all the image crops (a bag-of-words approach). We showed that these learned representations are more robust than baseline representations, which are based on the entire image. Moreover, choosing a value of \( n \) larger than two accelerates the learning of co-occurrence patterns between crops. In other words, CF-AMC-SSL learns meaningful and robust representations against adversarial attacks while converging with significantly fewer epochs.

Our work further verifies that free adversarial training can be effectively applied to self-supervised learning, irrespective of the specific loss function used, and remains effective even with a significant reduction in the number of training epochs.

From an adversarial robustness perspective, CF-AMC-SSL utilizes a comprehensive training objective that incorporates both a regularization term \( R(Z_i^{\text{adv}}) \) and an invariance term \( D(Z_i^{\text{adv}}, \bar{Z}_{\text{adv}}) \). The regularization term enhances the model's robustness by enforcing constraints on adversarial representations, while the invariance term promotes smoothness by ensuring that adversarial examples are aligned with their average representation \( \bar{Z}_{\text{adv}} \). This averaging mechanism effectively mitigates the impact of extreme adversarial perturbations by clustering adversarial examples around their average, leading to more stable and generalized representations. As a result, the model achieves a superior balance between adversarial robustness and clean accuracy, handling adversarial perturbations more effectively while maintaining high performance on clean data.

\section{Conclusion}

In conclusion, we explored the robustness of  Extreme-Multi-Patch Self-Supervised
Learning (EMP-SSL) against adversarial attacks using both standard and adversarial training techniques. Our findings underscored the significant impact of multi-scale crops within the robust EMP-SSL algorithm, enhancing model robustness without sacrificing accuracy. This improvement contrasts with robust SimCLR, which relies on only a pair of crops per image and necessitates more training epochs. Moreover, we demonstrated the efficacy of incorporating free adversarial training into methods like SimCLR and EMP-SSL, even though training epochs are limited in EMP-SSL. This integration resulted in the development of Cost-Free Adversarial Multi-Crop Self-Supervised Learning (CF-AMC-SSL), achieving substantial advancements in both robustness and accuracy while reducing training time. In summary, our study contributes to the advancement of self-supervised learning, making it more practical and impactful for real-world applications.


% \ifCLASSOPTIONcompsoc
  %The Computer Society usually uses the plural form
%   \section*{Acknowledgments}
% \else
%   % regular IEEE prefers the singular form
% \fi
\section*{Acknowledgment}
This work was partially supported by the National Science Foundation (Awards 2007202, 2107463, 2038080, and 2233873).



%\bibliographystyle{IEEEtran}

\appendix

\begin{figure*}[htbp]
    \centering
    \includegraphics[width=\textwidth]{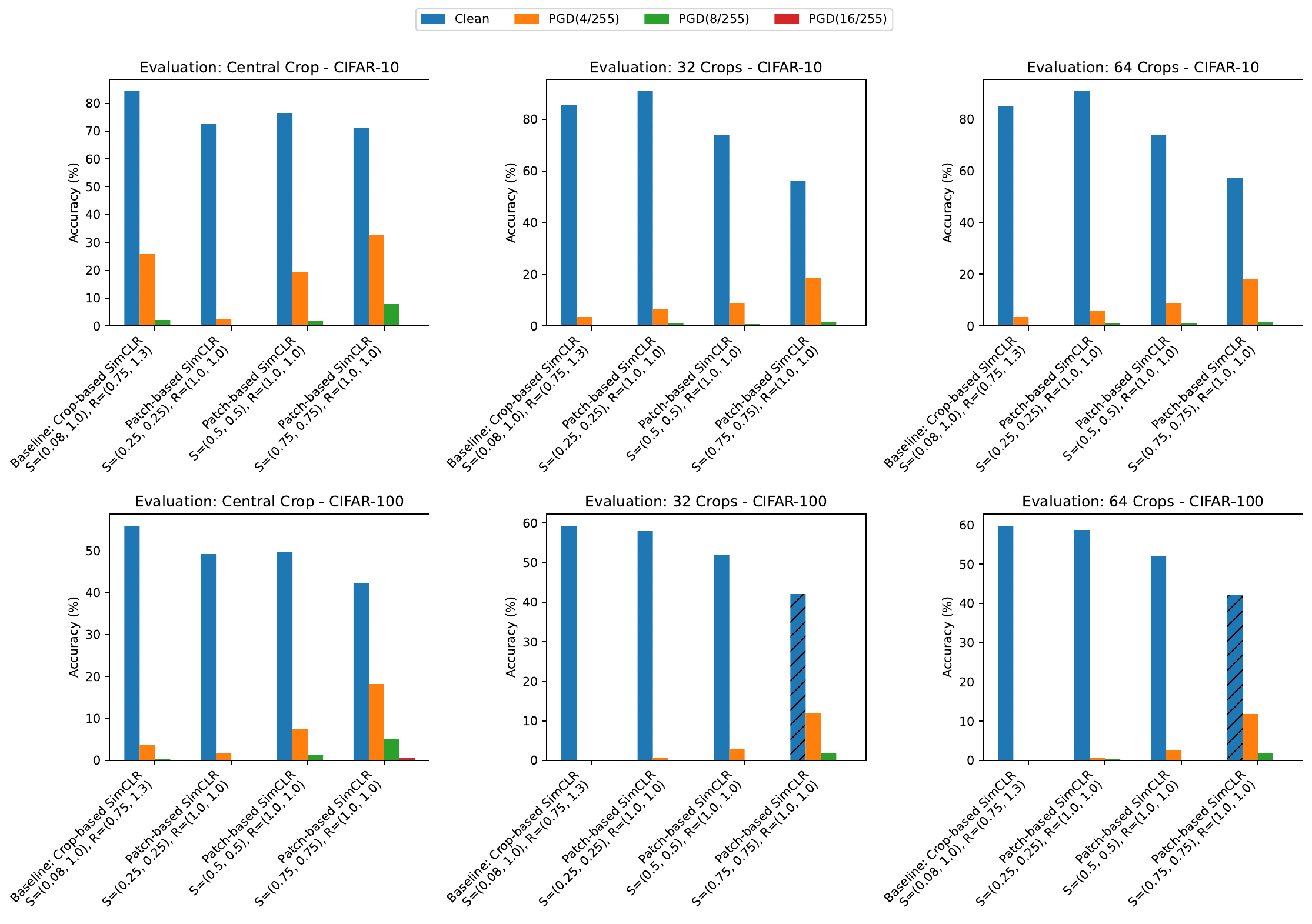}  % Replace with your PDF file name
    \caption{\textbf{Evaluating the robustness against PGD attacks through adversarial pretraining on CIFAR-10 and CIFAR-100 datasets, we compare the performance of patch-based SimCLR (with various patch sizes) to that of baseline SimCLR.} Our findings reveal a noticeable trade-off between clean accuracy and robustness. In addition, central cropping (first column) demonstrates higher efficiency in terms of overall complexity, clean accuracy, and robustness. Moreover, increasing patch sizes reduces clean accuracy but improves model robustness.  Note that the variables $S$ and $R$ correspond to the scales and ratios employed in the PyTorch framework's RandomResizedCrop method.}
    \label{eval-simclr}
\end{figure*}

\begin{figure*}[htbp]
    \centering
    \includegraphics[width=\textwidth]{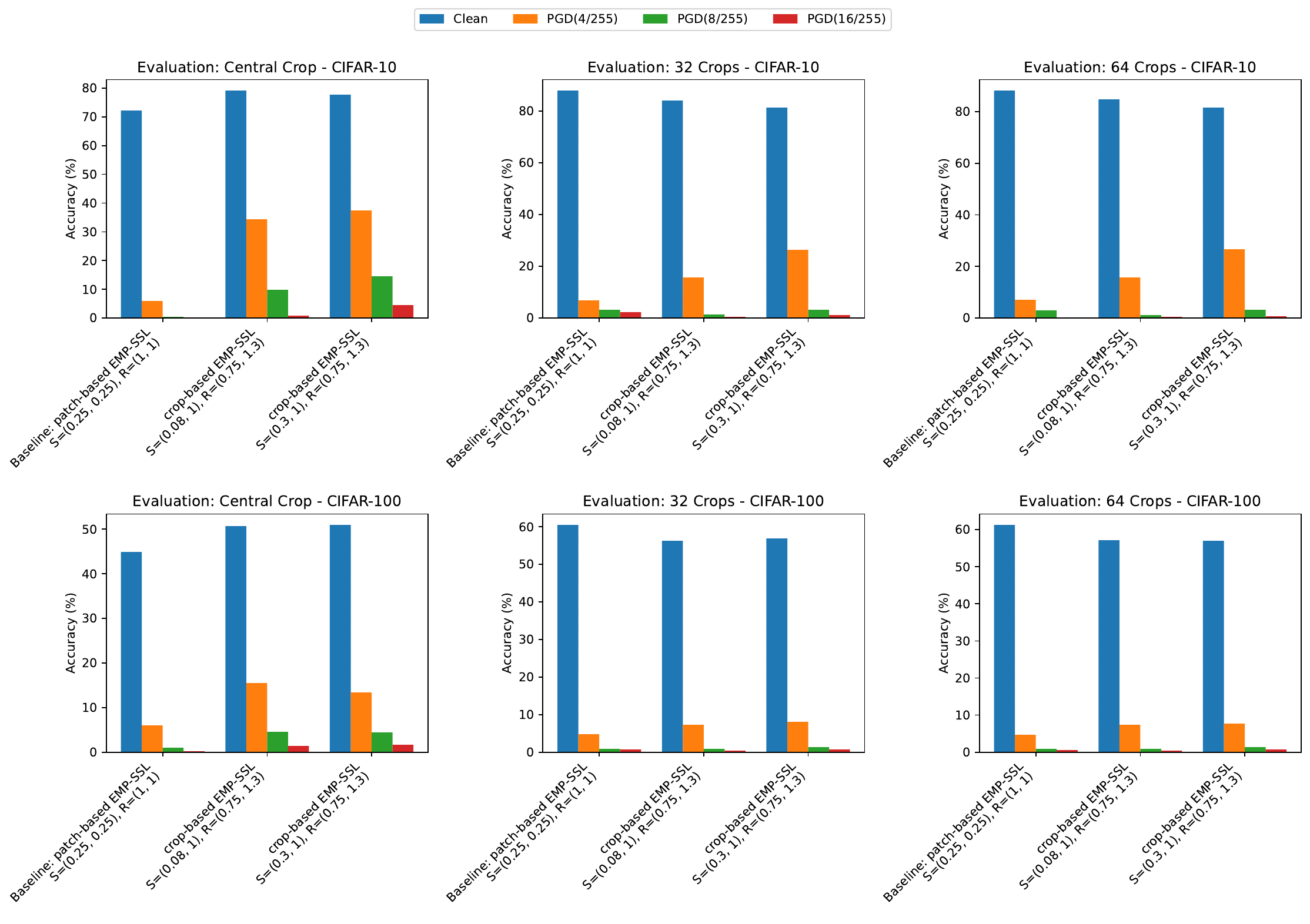}  % Replace with your PDF file name
    \caption{\textbf{Evaluating the robustness against PGD attacks through adversarial pretraining on CIFAR-10 and CIFAR-100 datasets, we compare the performance of crop-based EMP-SSL (with various crop sizes) to that of baseline EMP-SSL.} Our analysis reveals that the crop-based approach in EMP-SSL demonstrates enhanced robustness. Compared to the results presented in Figure \ref{eval-simclr}, it is clear that Robust EMP-SSL achieves a superior balance between clean accuracy and robustness, in contrast to robust SimCLR. Here, the variables $s$ and $r$ denote the scales and ratios utilized for the RandomResizedCrop method within the PyTorch framework.}
    \label{eval-empssl}
\end{figure*}

\subsection{Evaluation with 32- and 64-Patch Aggregation}

In addition to central cropping, we evaluated the robust base encoders using multi-patch aggregation with 32 and 64 patches. These methods involve aggregating embeddings from multiple fixed-size patches during evaluation. While this approach provides insights into the robustness of the learned representations, it is computationally more intensive than central cropping.

Key observations from the results (Figures \ref{eval-simclr} and \ref{eval-empssl}) are as follows:
\begin{itemize}
    \item \textit{Multi-patch aggregation enhances clean accuracy, especially when a larger number of patches (e.g., 64) is used.}
    \item \textit{However, the computational cost increases significantly, making central cropping more practical for resource-constrained settings.}
\end{itemize}

\subsection{Detailed Ablation Study of Robust EMP-SSL}

\begin{figure*}[htbp]
    \centering
    \includegraphics[width=0.87\textwidth]{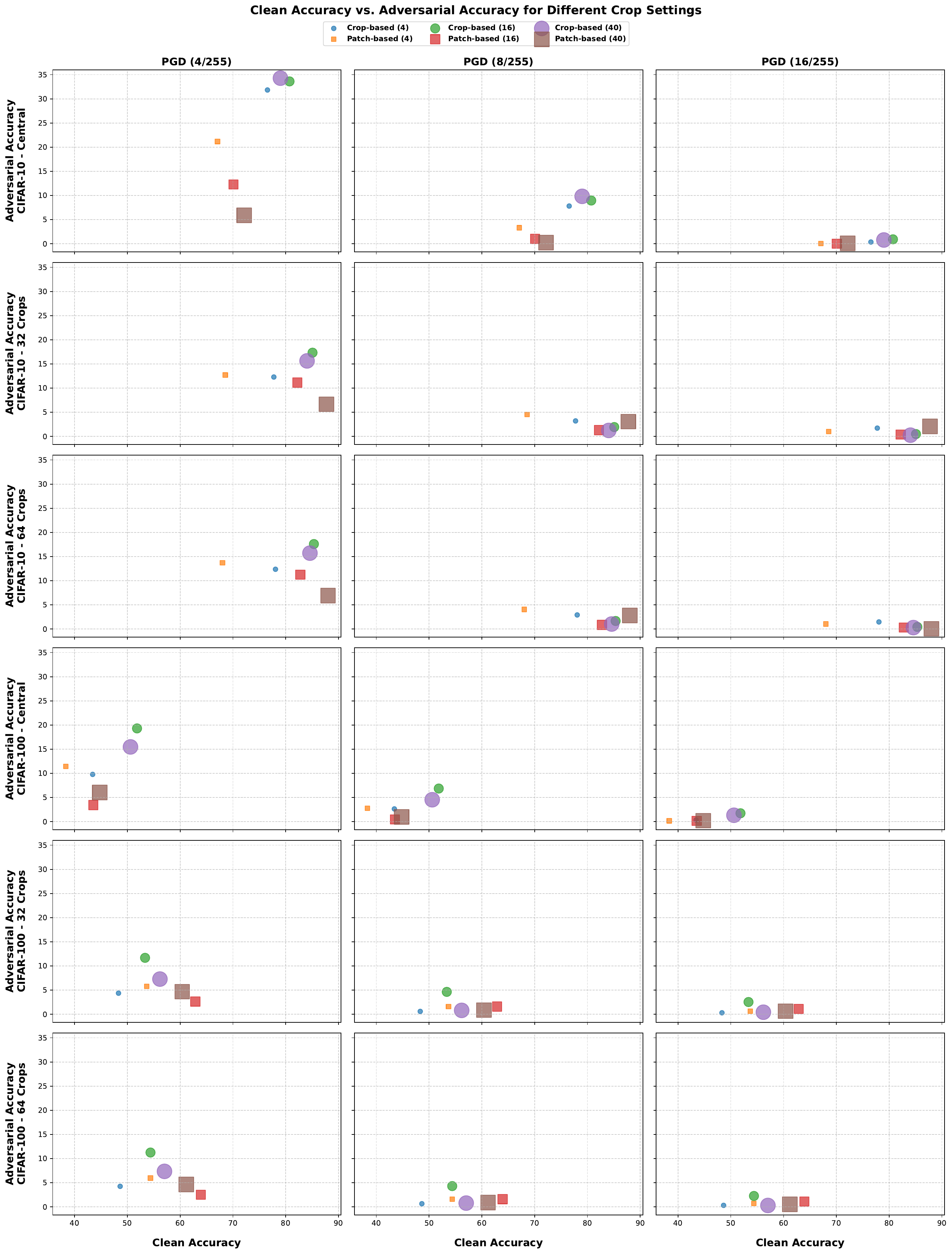}  % Replace with your PDF file name
    \caption{\textbf{Evaluation of robust EMP-SSL across different patch (crop) sizes on CIFAR-10 and CIFAR-100 datasets:} Our results emphasize that, when employing the patch-based EMP-SSL method with multi-patch aggregation during evaluation, a significant augmentation in the number of patches leads to a noticeable enhancement in clean accuracy. Furthermore, when using crop-based EMP-SSL with central-crop assessment, a more equitable balance between clean accuracy and model robustness can be established, especially evident with a moderate number of crops, such as 16. Note that "Crop-based (4)" means augmentation with scales (S) of (0.08, 1.0) and ratios (R) of (0.75, 1.3), with (4) denoting the number of crops. Similarly, "Patch-based (4)" involves scales (S) of (0.25, 0.25) and ratios (R) of (1.0, 1.0), with (4) representing the number of patches.}
    \label{eval-empssl-diff-patchsize}
\end{figure*}

The ablation study analyzed the impact of varying the number of patches (crops) used for adversarial training in the EMP-SSL framework. The results for CIFAR10 and CIFAR100 are shown in Figure \ref{eval-empssl-diff-patchsize}.

Findings include:
\begin{itemize}
    \item \textit{Increasing the number of patches during evaluation consistently improves clean accuracy in patch-based EMP-SSL.}
    \item \textit{Moderate numbers of crops (e.g., 16) with crop-based EMP-SSL preserve a better trade-off between clean accuracy and robustness when evaluated using central cropping.}
\end{itemize}

\end{document}